\documentclass[letterpaper]{article} 
\usepackage{aaai2026}  
\usepackage{times}  
\usepackage{helvet}  
\usepackage{courier}  
\usepackage[hyphens]{url}  
\usepackage{graphicx} 
\usepackage{natbib}  
\usepackage{caption} 
\usepackage{algorithm}
\usepackage{algorithmic}

\usepackage{newfloat}
\usepackage{listings}
\DeclareCaptionStyle{ruled}{labelfont=normalfont,labelsep=colon,strut=off} 
\floatstyle{ruled}
\newfloat{listing}{tb}{lst}{}
\floatname{listing}{Listing}
\title{EduMod-LLM: A Modular Approach for Designing Flexible and Transparent Educational Assistants}

\author{
    Meenakshi Mittal\textsuperscript{\rm 1},
    Rishi Khare\textsuperscript{\rm 1},
    Mihran Miroyan\textsuperscript{\rm 1},
    Chancharik Mitra\textsuperscript{\rm 2},
    Narges Norouzi\textsuperscript{\rm 1}
}
\affiliations{
    \textsuperscript{\rm 1}University of California, Berkeley\\
    \textsuperscript{\rm 2}Carnegie Mellon University \\
    meenakshi.mittal@berkeley.edu,
    rishi.khare@berkeley.edu, miroyan.mihran@berkeley.edu,
    cmitra@cs.cmu.edu,
    norouzi@berkeley.edu
}

\usepackage{bibentry}

\newcommand{\edisonmodel}{Edison}
\newcommand{\model}{EduMod-LLM}

\newcommand{\university}{University of California, Berkeley}
\usepackage{latexsym}
\usepackage{booktabs}

\usepackage{tcolorbox}

\usepackage{times}

\usepackage{graphicx}

\begin{document}

\maketitle

\begin{abstract}
With the growing use of Large Language Model (LLM)-based Question-Answering (QA) systems in education, it is critical to evaluate their performance across individual pipeline components. In this work, we 
introduce {\model}, a modular function-calling LLM pipeline, and present a comprehensive evaluation along three key axes:
function calling strategies, retrieval methods, and generative language models. 
Our framework enables fine-grained analysis by isolating and assessing each component. We benchmark function-calling performance across LLMs, compare our novel structure-aware retrieval method to vector-based and LLM-scoring baselines, and evaluate various LLMs for response synthesis. This modular approach reveals specific failure modes and performance patterns, supporting the development of interpretable and effective educational QA systems.
Our findings demonstrate the value of modular function calling in improving system transparency and pedagogical alignment. Website and Supplementary Material: \url{https://chancharikmitra.github.io/EduMod-LLM-website/}

\end{abstract}

\section{Introduction}

Modern LLMs have demonstrated impressive capabilities across a range of Natural Language Processing (NLP) tasks, including question-answering, knowledge-base retrieval, and summarization. A variety of methods and architectures have advanced progress in these areas including Chain-of-Thought (CoT) prompting~\cite{wei2022chainofthought}, Reinforcement Learning (RL) for reasoning~\cite{guo2025deepseek, OpenAI2024O1SystemCard}, RAG~\cite{lewis2020retrievalaugmented, karpukhin2020dense}, robust function calling capabilities~\cite{OpenAI2023FunctionCalling,Lu2023ChameleonPC, Schick2023ToolformerLM}, and LLM token probability scoring for retrieval and automated evaluation~\cite{lin2024vqascore, zheng2023judging}. However, it is unclear how these advances translate to real-world educational settings, where questions often require alignment with instructional goals, compliance with course policies, and sensitivity to student learning trajectories.

While LLMs have achieved impressive results on standardized benchmarks, such as solving complex mathematical problems in American Mathematics Competitions (AMC) and American Invitational Mathematics Examination (AIME) competitions~\cite{lewkowycz2022solving, trinh2023solving}, their effectiveness in a live classroom environment is uncertain. 
Classroom questions are rarely clearly structured; they are often ambiguous, contextually grounded, and embedded in course-specific terminology. 
Similarly, retrieval paradigms have advanced from basic encoder-based embedding approaches~\cite{reimers-2019-sentence-bert} to fine-tuned generative models~\cite{luo2024nv, wang2023improving} that achieve strong performance on benchmarks like Massive Text Embedding Benchmark (MTEB)~\cite{muennighoff2022mteb}. Yet, these methods often struggle with the heterogeneous nature of educational data, ranging from textbooks, assignments, student Q\&A, and course logistics. Moreover, evaluation techniques such as token probability scoring~\cite{lin2024vqascore, zheng2023judging}, although promising, remain unsuitable for educational use cases where pedagogical alignment, e.g., avoiding direct answers during assessments and adhering to instructional norms of the course, is as critical as factual correctness~\cite{dorodchi2019authentic, denny2024computing}. Existing benchmarks fail to capture these instructional subtleties and leave a gap in understanding how LLM components perform under real educational constraints.

To address this gap, we introduce \textbf{{\model}}, a \textbf{modular LLM-based framework} for student QA, designed to evaluate the individual and combined impact of three critical components: \textit{function calling strategies}, \textit{retrieval methods}, and \textit{generative language models}. This modular design improves transparency by enabling the precise diagnosis of failure modes and performance bottlenecks using real student questions from a large-scale university course. By systematically isolating and analyzing each module, we provide new insights into how to build more interpretable and pedagogically-aligned educational QA systems.

Our contributions in this work are threefold:
\begin{enumerate}
\item \textbf{LLM-as-a-Judge Evaluation Framework:} We design and validate an LLM-based evaluation module aligned with experienced Teaching Assistants (TAs). This approach provides scalable and automated assessments of response quality, factual accuracy, and pedagogical appropriateness to student questions. This evaluation represents a first-of-its-kind assessment of \textit{LLMs and Large Reasoning Models (LRMs) as reward models for course-specific educational QA}, distinguishing itself from general reward-model benchmarks~\cite{Li2024VLRewardBenchAC, Yasunaga2025MultimodalRH, Li2024VLRewardBenchAC} 
\item \textbf{Modular Pipeline Evaluation:} We develop a novel modular LLM pipeline that allows each component—function calling, retrieval, and response generation-to be independently and jointly evaluated. This enables fine-grained comparisons and targeted improvements in performance, establishing how modularity is especially important in educational QA.
\item \textbf{Structure-Aware Retrieval:} We introduce a novel retrieval method that reflects the hierarchical organization of educational content. By summarizing and indexing materials, such as textbooks and assignments, according to their natural structure (chapters, sections, and problem numbers), our method improves both retrieval precision and interpretability.

\end{enumerate}
Our experimental results reveal several key findings. Firstly, our LLM-as-a-Judge results suggest that non-reasoning models in particular such as DeepSeek-V3~\cite{DeepSeekAI2024DeepSeekV3TR} can reliably mimic TA evaluation standards, enabling scalable and expert-aligned quality control. Secondly, we find that our novel multi-step function-calling approaches match and outperform SoTA methods, even those with TA-designed function-calling rules. Secondly, we find that our novel structure-aware retrieval and carefully selecting GPT-4.1 as the base LLM contributes significantly to better responses, indicating the importance of specializing retrieval methods and LLMs for educational QA tasks. 
Collectively, these findings offer actionable design principles for building robust, transparent, and instructionally sound AI-ED systems.

\begin{figure*}[ht]
    \centering
    \includegraphics[width=.9\linewidth]{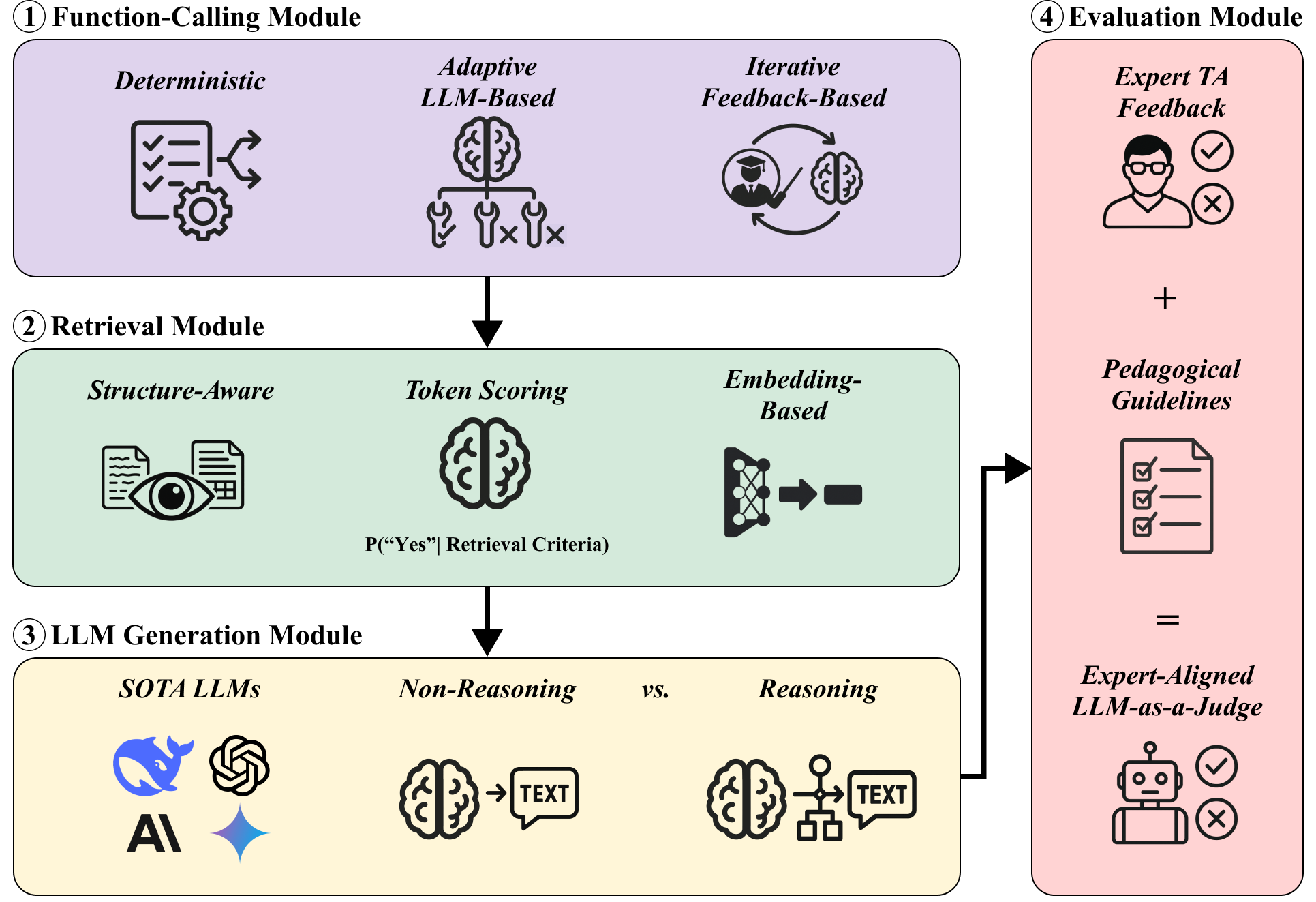}
    \caption{\textbf{\model} {\model} is a modular approach to developing an LLM-based pipeline for answering student questions. We explore design considerations in function-calling, retrieval, and LLM response generation via a high-quality and scalable evaluation module that leverages expert TA insights. This approach enables flexibility to courses and technical constraints as well as transparency for iterative improvement.}
    \label{fig:main_figure}
\end{figure*}

\section{Related Work}

\subsection{LLM-powered Virtual Assistants in Education}

Large classroom settings often suffer from logistical inefficiencies and repeated questions, which creates a strong use case for integrating LLM into educational workflows~\cite{malan_cs50,mirhosseini_what_2023, ZamfirescuPereira202461ABotAH, mitra2024retllm, codeaid}. Research shows that student motivation is correlated with their help-seeking behaviors~\cite{Oberman2004MotivationAA}. Digital tools can help bridge this gap~\cite{Ali2021UsingPA, CampilloFerrer2020GamificationIH}, making LLMs such as GPT-4~\cite{Achiam2023GPT4TR}, LLaMA~\cite{touvron2023llama}, and Gemini~\cite{Reid2024Gemini1U} promising candidates for educational assistance~\cite{Gan2023LargeLM, codeaid, ZamfirescuPereira202461ABotAH, CS50, mitra2024retllm, Wang2024LargeLM}. These approaches are often unified pipelines that lack modularity and transparency necessary for iterative improvement. Educational contexts impose unique demands: responses must be factually correct, pedagogically appropriate, interpretable by instructors, and aligned with instructional policies. These constraints challenge black-box LLMs and require transparent and controllable QA systems for classroom use.

\subsection{Transparent LLM Pipelines}

As LLMs become more complex, understanding and controlling their behavior become a central concern. Traditional explainability methods, such as feature attribution techniques~\cite{sundararajan2017axiomatic,lundberg2017unified}, are difficult to scale to models with billions of parameters. More recent techniques offer structural insights using task vectors or a sparse attention mechanism to attribute latent components to high-level behavior~\cite{hendel2023task,todd2023function,huang2024sparse,mitra2024interpretable}. 
Complementary approaches in natural language include generating self-explanations or summaries~\cite{huang2023can,zhao2023explainability}, as well as prompting strategies such as CoT prompting and In-Context Learning (ICL)~\cite{Wei2022ChainOT, Brown2020LanguageMA}. These methods improve transparency, but few have been systematically evaluated in educational settings.

Our work builds on this line of research by focusing on the faithfulness~\cite{jacovi2020annotating} and plausibility~\cite{shen2022plausible} of the output of the LLM models in the specific context of student question answering. We extend this literature by designing a modular pipeline that is both interpretable and aligned with the instructor's expectations.

\section{Methods}
{\model} is a modular framework for student question-answering to analyze the strengths and weaknesses of different pipeline components systematically. By isolating the \textbf{functional-calling}, \textbf{retrieval}, and \textbf{answer generation} modules, our framework (shown in Figure~\ref{fig:main_figure} enables detailed analysis of where failure occurs and how each component contributes to the final response quality. In the following, we describe the dataset (subsection~\ref{sec:data}), the architecture of each module (subsection~\ref{sec:fc_pipeline}, Section~\ref{sec:retrieval}, and subsection~\ref{sec:llmjudge}), and our automated pedagogical quality evaluation framework (subsection~\ref{sec:llmjudge}). Synthetic examples of EduMod-LLM's TA-in-the-Loop feedback incorporation is shown in Supplementary Section~\ref{appendix:sec:synthetic_examples}.

\subsection{Data}
\label{sec:data}
Evaluating our approach on real student questions asked in a course environment is a key factor in assessing the effectiveness of different SOTA LLM-based approaches in education. We collected historical student EdSTEM data from an upper-division undergraduate data science course at an R-1 institution in the USA, during the Spring 2024 semester. The course served 1133 undergraduates and 71 graduate students, supported by 59 staff members, covering topics ranging from data processing to machine learning and probabilistic modeling. Ground truth responses to these questions are generated via a recent SOTA, TA-in-the-loop educational assistant {\edisonmodel}, which was integrated into the course's EdSTEM\footnote{\url{https://edstem.org/}} discussion forum. 

\paragraph{Target Course.}
The dataset was collected from DATA 100 (Principles and Techniques of
Data Science), a single upper-division undergraduate data science course at the University of California, Berkeley. 
The course spans 17 weeks and covers topics from data processing to machine learning and probabilistic modeling. 
It includes approximately 11 weekly programming assignments, 13 lab exercises, and 2 exams (midterm and final). 

Students use the EdSTEM discussion forum for Q\&A related to conceptual understanding, assignments,  and logistics throughout the semester. To clarify briefly, conceptual questions are those related to course concepts and not directly referencing assignments. Logistics questions relate to those about deadlines, course resources, and generally any other course policies.

When students submitted new questions, they could opt in to receiving assistance from {\edisonmodel}. TAs reviewed the questions and, when appropriate, triggered {\edisonmodel} to generate draft responses, which TAs could then edit and post. 
Our final dataset contains 1,000 student-consented questions, selected randomly from the forum, spanning the full 17 weeks to ensure temporal diversity across assignments and topics. We also collected their corresponding TA-written responses.

\subsection{Function-Calling Pipeline}
\label{sec:fc_pipeline}

Educational QA requires access to multiple information sources, such as assignments, textbooks, logistics documents, and past student Q\&A. Prior work has implemented RAG using these sources~\cite{mitra2024retllm, Miroyan2025AnalyzingPQ}. We reformulate each source-specific retrieval strategy as a \textbf{function} to allow LLM to dynamically decide which to call during response generation. We provide the LLM with the following function:

\begin{itemize}
    \item \texttt{qa\_retrieval(query, top\_k)}: Retrieves \texttt{top\_k} similar Q\&A pairs from prior semesters' EdSTEM forums.
    \item \texttt{textbook\_retrieval(query)}: Retrieves relevant chunks from the textbook and course notes.
    \item \texttt{assignment\_retrieval(query)}: Retrieves relevant portions of homework, labs, and projects (including any relevant solutions).
    \item \texttt{logistics\_retrieval(query)}: Retrieves matching content from course logistics (e.g., syllabus, policies).
\end{itemize}

The LLM uses OpenAI's Function-calling (FC) API \cite{openaifc} to choose functions, generate input arguments (e.g., \texttt{query}), and uses the returned outputs when composing its own response. 
Unless otherwise specified, the model can call zero or more functions in each turn. 
Prompts are included in Appendix Section~\ref{supp: additional_impl_details}. For all the experiments, any student questions containing images (e.g., code screenshots) are pre-processed using Optical Character Recognition (OCR).

We implement and compare the following pipelines:

\begin{itemize}
    \item \textbf{Baseline} (\texttt{Edison}): A SoTA deterministic function-calling pipeline~\cite{Miroyan2025AnalyzingPQ} that requires TA hand-designed rules based on course-specific metadata.
    
    \item \textbf{Pipeline 1} (\texttt{fc}): A default FC baseline with optional function invocation.
    
    \item \textbf{Pipeline 2} (\texttt{fc\_categorize}): Prepends a student-selected category to guide the LLM to perform function selection.
    
    \item \textbf{Pipeline 3} (\texttt{fc\_forced}): Requires the model to select at least one function.
    
    \item \textbf{Pipeline 4} (\texttt{fc\_iterative}): Requires the model to select at least one function, then allows further function calls based on retrieved content.

    \item \textbf{Pipeline 5} (\texttt{fc\_feedback}): After an initial function call (required) and response generation, the output is reviewed by a judging model (LLM-as-a-Judge), which provides feedback. The model then re-enters the FC loop with this feedback.
    
    \item \textbf{Pipeline 6} (\texttt{fc\_multihop}):
    Decomposes function selection and argument generation into two LLM calls to simulate a multi-hop reasoning process.


\end{itemize}

We evaluate these pipelines along two axes: \textbf{function selection accuracy} (compared to expert-labeled ground-truth) and \textbf{final response quality} (via human and LLM-based evaluation).

\subsection{Structure-Aware Retrieval}
\label{sec:retrieval}
We introduce an interpretable hierarchical retrieval pipeline designed to extract logically coherent question-and-answer pairs from raw assignment documents. This approach addresses two key challenges: (1) eliminating the overhead and scalability limitations of requiring instructors to manually extract and upload individual questions and answers, and (2) improving retrieval quality beyond vector-based methods, which often operate on shallow and fixed-length chunks.

\paragraph{Chunking}
We split documents into fixed-length chunks. GPT-4o is used to detect question headers and generate a complete list of question boundaries while removing any extraneous information. These headers are then used to segment the document into coherent question-and-answer blocks.

\paragraph{Retrieval}
Next, we recursively summarize the chunks into a hierarchical (k-ary tree) structure. Each parent node summarizes its children, enabling coarse-to-fine traversal.  
At inference time, the LLM selects top-level summaries from a table of contents, then performs a beam-k search through the document tree to locate the most relevant content (chunk).
This design improves both the relevance of the retrieved materials and the interpretability of the retrieval module, making it possible to trace which material was used and why.

\subsection{Expert TA-aligned LLM-as-a-Judge Module:}
\label{sec:llmjudge}
To evaluate generated answers at scale, we implement an LLM-based evaluation module aligned with expert TA grading standards.
Building on the taxonomy from prior work \cite{Miroyan2025AnalyzingPQ}, we collaborated with two expert TAs to define detailed rubrics for three key dimensions. 
\begin{figure*}
    \centering
    \includegraphics[width=1\linewidth]{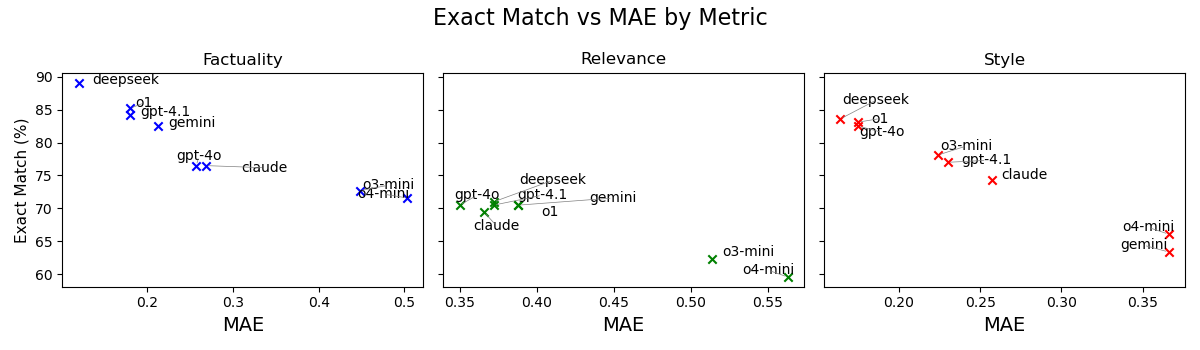}
    \caption{\textbf{Exact Match vs. MAE for LLM-as-a-Judge model across factuality, relevance, and style.} DeepSeek generally achieves the best alignment with TA responses when measured across exact match and MAE.}
    \label{fig:llm_judge}
\end{figure*}

\begin{figure}[ht]
    \centering
    \includegraphics[width=1\linewidth]{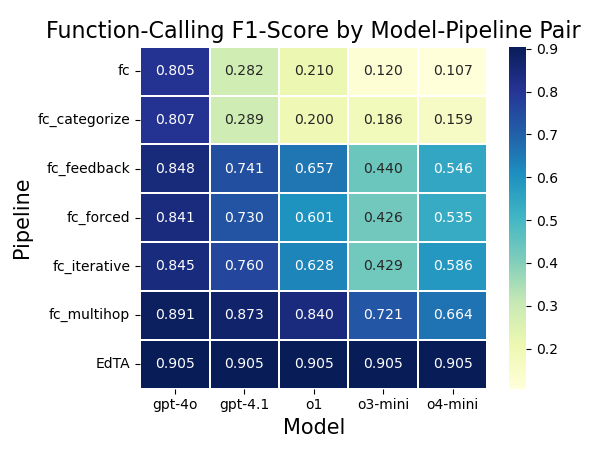}
    \caption{\textbf{Function-calling F1 Scores by Model and Pipeline.} GPT-4o and GPT-4.1 achieve the highest accuracy across all models, while \texttt{fc\_multihop} is the best . The rule-based \texttt{Edison} does not rely on LLM function-calling and thus achieves the same score for each model.}
    \label{fig:function_call}
\end{figure}
\begin{tcolorbox}[title=Evaluation Criteria]
\textbf{Factuality:} Assesses the accuracy of the information.

\textbf{Relevance:} Evaluates alignment with (1) course material and (2) the specific student question.

\textbf{Style:} Examines (1) clarity, (2) appropriate verbosity (concise for assignments, detailed for concepts), and (3) absence of direct solutions.
\end{tcolorbox}

We encode these criteria into few-shot prompts that are used by multiple LLMs to simulate expert assessment. Our validation shows a high agreement between these LLM judges and the TA scores, supporting their use for large-scale evaluation while preserving pedagogical alignment.

\section{Experiments and Results}

We evaluate our modular LLM pipeline along three key axes: \textbf{function selection accuracy} in Subsection~\ref{subsec:func_call_eval}; \textbf{retrieval performance} in Subsection~\ref{subsec:retrieval_eval}; and \textbf{response quality} in Subsection~\ref{subsec:llm_response_eval}. Across all experiments, we use a consistent set of metrics:
\begin{itemize}
    \item For FC, we report \textbf{F1 score} against expert-annotated function labels.
    \item For QA response quality, we report \textbf{Likert-scale scores} for \textbf{factuality} (1-5), \textbf{relevance} (1-5), and \textbf{style} (1-3), using both expert TA labels and an LLM-as-a-Judge module.
    \item For retrieval, we report \textbf{Recall@k} for k$=\{1, 3, 5\}$ for document-level relevance.
\end{itemize}

A subset of 180 out of 1,000 student questions was manually labeled by two expert TAs for answer quality evaluation and function selection. Appendix Section~\ref{appendix:sec:human_eval_details} contains more details about our human evaluation.


\subsection{LLM-as-a-Judge Evaluation}
\label{subsec:llm_judge_eval}
To scale evaluation beyond manual annotations, we implement an LLM-as-a-judge module aligned with TA preferences. We collaborated with two expert TAs to develop a refined rubric for factuality, relevance, and pedagogical style, and embedded these criteria into few-shot prompts for consistent scoring.

Our expert-labeled validation set contained 180 samples, selected to balance coverage across topics and question types. 
While this relatively small sample provides high-quality human judgments, it may not capture the full diversity of 1{,}000 questions. 

To mitigate sampling effects, we used five random seeds for model runs and averaged results with 95\% confidence intervals, 
which reduced variance across evaluations. 
Future extensions will increase the annotated subset and explore active-learning–based selection to ensure broader coverage.


We compare various LLMs in their ability to serve as judges, using two metrics: \textbf{Exact Match} with TA scores and \textbf{Mean Absolute Error (MAE)}. 
As shown in Figure~\ref{fig:llm_judge},
DeepSeek-v3 achieves the best alignment with expert TAs, with over 70\% exact match across all categories (close to 90\% for factuality) and the lowest MAE across all categories. This consistent performance across all dimensions makes DeepSeek-v3 particularly suitable for the multifaceted evaluation required in educational contexts, where responses must be simultaneously factual, relevant, and stylistically appropriate.

Other models (GPT-4o, Claude, Gemini, and o3/o4-mini) exhibit partial strength but do not match DeepSeek's consistency.
This confirms that LLM-based evaluation, when guided by a domain-specific rubric, can serve as a scalable and accurate proxy for assessment in educational contexts.

\subsection{Function-Calling Evaluation}
\label{subsec:func_call_eval}
We assess the accuracy of FC pipelines by comparing their selected functions to ground-truth labels established by two expert TAs. Each experiment is run five times to account for LLM variability, and we compute the mean F1 score for function selection.


As shown in Figure~\ref{fig:function_call}, GPT-4o consistently achieves the highest accuracy across all pipelines, with F1 scores exceeding 0.80 in most cases. More advanced pipelines such as \texttt{fc\_multihop} outperform simpler methods across all models, which confirms the value of the use of iterative and feedback-aware tool usage. Most importantly, the deterministic \texttt{Edison} pipeline performs comparatively (F1-0.905). This suggests that in structured educational settings, flexible LLM-based function selection can outperform TA-designed, rule-based function selection.

\label{subsec:retrieval_eval}
\begin{table*}[htbp]
   \centering
   \begin{tabular}{lccc}
       \toprule
       \textbf{Method} & \textbf{Recall@1} & \textbf{Recall@3} & \textbf{Recall@5} \\
       \midrule
       gemini-embedding-exp-03-07 & 0.441 $\pm$ 0.091 & 0.487 $\pm$ 0.091 & 0.523 $\pm$ 0.091 \\
       \addlinespace
       text-embedding-ada-002 & 0.460 $\pm$ 0.091 & 0.640 $\pm$ 0.090 & 0.721 $\pm$ 0.086 \\
       \addlinespace
       vector\_gen & 0.451 $\pm$ 0.091 & 0.676 $\pm$ 0.086 & 0.757 $\pm$ 0.081 \\
       \addlinespace
       \midrule
       \textbf{hier\_gen} & \textbf{0.784 $\pm$ 0.077} & \textbf{0.874 $\pm$ 0.059} & \textbf{0.946 $\pm$ 0.041} \\
       \bottomrule
   \end{tabular}
      \caption{\textbf{Retrieval Performance Across Different Methods}. Recall@1, 3, and 5 are reported with 95\% confidence intervals.}
      \label{tab:retrieval}
\end{table*}

\begin{table*}[htbp]
    \centering
    \begin{tabular}{llcccc}
        \toprule
        \textbf{Model} & \textbf{Experiment} & \textbf{Factuality} & \textbf{Relevance} & \textbf{Style} & \textbf{FC F1 Score} \\
        \midrule
        GPT-4o & fc & 4.508 $\pm$ 0.147 & 4.902 $\pm$ 0.091 & 2.918 $\pm$ 0.074 & 0.8055 \\
        \addlinespace
        GPT-4o & fc\_categorize & 4.541 $\pm$ 0.156 & 4.902 $\pm$ 0.091 & 2.934 $\pm$ 0.057 & 0.8066 \\
        \addlinespace
        GPT-4o & fc\_feedback & 4.508 $\pm$ 0.156 & 4.885 $\pm$ 0.074 & 2.902 $\pm$ 0.074 & 0.8481 \\
        \addlinespace
        GPT-4o & fc\_forced & 4.574 $\pm$ 0.147 & 4.885 $\pm$ 0.115 & 2.967 $\pm$ 0.041 & 0.8410 \\
        \addlinespace
        GPT-4o & fc\_iterative & 4.623 $\pm$ 0.131 & 4.918 $\pm$ 0.074 & 2.951 $\pm$ 0.058 & 0.8454 \\
        \addlinespace
        GPT-4o & Edison* & 4.541 $\pm$ 0.164 & 4.918 $\pm$ 0.082 & 2.934 $\pm$ 0.057 & \textbf{0.9049} \\
        \addlinespace
        GPT-4o & fc\_multihop & \textbf{4.689 $\pm$ 0.115} & \textbf{4.951 $\pm$ 0.058} & \textbf{2.951 $\pm$ 0.058} & 0.8913 \\
        
        \bottomrule
    \end{tabular}
        \caption{\textbf{Evaluation of FC Pipelines}. For all FC pipelines GPT-4o is used as the base LLM and the quality of generated responses are evaluated across multiple axes with 95\% confidence intervals. FC F1 score refers to the F1 score performance of the function calling approach. *Note:  {\edisonmodel} requires an additional TA-designed, deterministic function-calling approach. }
        \label{tab:func_call_vs_quality}
\end{table*}

\subsection{Function-Calling and Response Quality}
To understand how function-calling quality relates to response quality, we analyzed the relationship between function selection accuracy and response metrics. Table~\ref{tab:func_call_vs_quality} presents a comparison of different function-calling approaches using GPT-4o (previously identified as the best-performing model for function calling) across our evaluation dimensions. The results demonstrate the impact of function-calling strategies on response quality metrics. Our automated \texttt{fc\_multihop} demonstrates slight improvements across all three metrics of factuality, relevance, and style, importantly, matching or excedding the performance of the \texttt{\edisonmodel} approach (using rule-based methods developed with TA insights).
This is particularly notable as the \texttt{fc\_multihop} approach does not require pre-determined question-category information from dataset metadata, which was necessary in creating the hand-designed function-calling in \texttt{\edisonmodel}. The \texttt{fc\_iterative} approach also shows strong performance, though not quite reaching the levels of \texttt{fc\_multihop}. These findings confirm that advanced LLM-based function-calling approaches can achieve comparable response quality to hand-designed rule-based methods (which leverage question metadata) while offering significantly greater flexibility as a result. Such flexibility enables more adaptable deployments across different courses and forums. 

\begin{table*}[ht]
   \centering
   \begin{tabular}{lrrr}
       \toprule
       \textbf{Model} & \textbf{Factuality} & \textbf{Relevance} & \textbf{Style} \\
       \midrule
       GPT-4o & 4.533 $\pm$ 0.040 & 4.930 $\pm$ 0.018 & 2.980 $\pm$ 0.010 \\
       O4-Mini-High & 4.574 $\pm$ 0.038 & 4.933 $\pm$ 0.020 & 2.985 $\pm$ 0.008 \\
       Claude-3.7-Sonnet & 4.551 $\pm$ 0.041 & 4.897 $\pm$ 0.024 & 2.995 $\pm$ 0.005 \\
       Gemini-1.5-Pro & 4.495 $\pm$ 0.041 & 4.920 $\pm$ 0.020 & 2.994 $\pm$ 0.005 \\
       DeepSeek-V3 & 4.586 $\pm$ 0.039 & 4.921 $\pm$ 0.021 & 2.997 $\pm$ 0.004 \\
       DeepSeek-R1 & 4.603 $\pm$ 0.037 & 4.944 $\pm$ 0.017 & 2.989 $\pm$ 0.007 \\
       \midrule
       Edison* (GPT-4o) & 4.553 $\pm$ 0.070 & 4.912 $\pm$ 0.039 & 2.997 $\pm$ 0.005 \\
       \textbf{GPT-4.1 (Ours)} & \textbf{4.648 $\pm$ 0.036} & \textbf{4.955 $\pm$ 0.014} & \textbf{2.999 $\pm$ 0.002} \\
       \bottomrule
   \end{tabular}
   \caption{\textbf{Evaluation of LLM Base Models} We evaluate across 1000 samples using \texttt{fc\_multihop} with GPT-4o function-calling. We report on our metrics scored using DeepSeek-V3 as a judge with 95\% confidence intervals. *Note:  {\edisonmodel} requires an additional TA-designed, deterministic function-calling approach. }
   \label{tab:model_performance}
\end{table*}

\subsection{Structure-Aware Retrieval Evaluation}

We evaluate different retrieval approaches for accessing course materials relevant to student questions. For this evaluation, we curated a set of 111 assignment-based student questions from course forums. To establish ground truth, we conducted human annotation to identify relevant documents for each question, treating unselected documents as negatives. This enabled us to calculate meaningful recall metrics for each retrieval method.

We compare to the following baselines:
\begin{itemize}

    \item \textbf{Vector Retrieval + Generative Ranking} (\texttt{vector-gen}): Vector-based retrieval, retrieving a fixed 15 chunks, then ranking these chunks using a generative scoring metric. This metric uses the probability score from applying softmax to the output logit of the token "yes" in the GPT-4o response to the question: "Is this document relevant for answering the student question?"~\cite{lin2024vqascore, zheng2023judging} before returning the top-k chunks. This probability is used directly as an alignment score for retrieval.
    
    \item \textbf{Our LLM-based Hierarchical Retrieval} (\texttt{hier\_gen}): The hierarchical generative retrieval method described earlier to retrieve the top-k chunks.
    
    \item \textbf{text-embedding-ada-001}: A vector-based retrieval pipeline based on Azure AI Search that performs hybrid search by combining vector similarity with keyword-based ranking, retrieving the top-k chunks.
    
    \item \textbf{gemini-embedding-exp-03-07}: A powerful embedding retrieval approach that leverages Gemini as a base model, enabling top-5 performance on the MTEB~\cite{muennighoff2022mteb} benchmark.
\end{itemize}

Table~\ref{tab:retrieval} presents the results of our evaluation. For each method, we computed recall rates at different cutoff thresholds (k=1, k=3, and k=5), measuring the proportion of relevant documents successfully retrieved in the top-k results.

Our structure-aware hierarchical retrieval approach (\texttt{hier\_gen}) substantially outperforms all compared methods across all recall metrics, achieving 0.784 Recall@1 and 0.946 Recall@5. This represents a significant, almost 30\% improvement over state-of-the-art embedding-based methods (\texttt{gemini-embedding-exp-03-07} and \texttt{text-embedding-ada-002}). These results underscore the significance of tailoring retrieval methods to the educational domain, where documents have diverse layouts and organizational structures. Despite the sophisticated representation learning in current SOTA embedding models, they still fall short when handling the unique challenges of educational content retrieval. By accounting for structural information in chunking and summarization, our approach achieves more effective retrieval, ultimately contributing to better response quality in educational QA systems. This finding suggests that educational retrieval represents a distinct problem domain requiring specialized approaches beyond general-purpose embedding models.

\subsection{LLM Response Generation Evaluation}
\label{subsec:llm_response_eval}
After identifying optimal FC, retrieval, and evaluation configurations, we evaluated which LLM performs best at response generation for student questions. Table~\ref{tab:model_performance} presents the performance of various models across our three evaluation dimensions, along with automated metrics. While all models perform reasonably well, GPT-4.1 stands out with superior scores across all dimensions, particularly in relevance (4.955) and style (2.999).


This result indicates that reasoning models don't necessarily surpass non-reasoning models in student QA retrieval-augmented generation, likely due to tradeoffs between better instruction-following/context utilization abilities of non-reasoning models compared to the deliberate thinking of reasoning models. The evaluation was successfully scaled to 1,000 questions using our LLM-as-a-Judge approach, demonstrating the value of tailored automated metrics for large-scale evaluation of educational AI assistants.

\section{Conclusion}

We present {\model} a modular function-calling framework to improve the interpretability of LLM-generated responses to student questions on educational course discussion boards. Through comprehensive evaluation of each component, we have gained significant insights into designing effective educational QA systems.

LLM-as-a-Judge experiments show that non-reasoning models can effectively follow TA evaluation guidelines to provide reliable assessments that correlate strongly with expert human judgments. This indicates that LLM-as-a-Judge approaches can scale fine-grained and domain-specific evaluation of the quality of responses to student questions. 

Our function-calling experiments reveal that correct function selection impacts response quality. Our multihop approach \texttt{fc\_multihop} with GPT-4o yields responses that match or exceed those outputted by the TA-designed \texttt{\edisonmodel} pipeline in relevance, factuality, and style, with less hand-engineering and course forum metadata to deploy. This leads to greater flexibility in its application to different courses.

Our structure-aware retrieval module demonstrates that domain-specific adaptations substantially outperform even top-performing general-purpose embedding models from the MTEB benchmark. This roughly 30\% improvement underscores that educational content retrieval represents a distinct problem domain where the hierarchical and structured nature of course documents demands specialized approaches beyond semantic similarity.

In LLM response generation, we found that GPT-4.1 surpassed the performance of most models (particularly GPT-4o and Gemini-1.5-Pro), including specialized reasoning models. This reveals that while LRMs show extraordinary performance on mathematics benchmarks, they may struggle to surpass LLMs in contextually-rich educational environments. Finally, we find that our additions of multihop function calling, structure-aware retrieval, and GPT-4.1 response generation help surpass the prior art (particularly in response factuality) {\edisonmodel}~\cite{Miroyan2025AnalyzingPQ}, which leverages deterministic function-calling, vector retrieval, and GPT-4o-based pipeline.

Collectively, our findings suggest that modularity can be a valuable principle when designing educational AI systems, allowing for flexible component-wise evaluation and adaptation to the specific demands of educational contexts. The modular approach enables transparent assessment of each component's contribution to overall system performance on real student questions, moving beyond the limitations of static benchmarks. Our experiments highlight the primary importance of structure-aware retrieval, with function-calling and LLM selection serving as important but secondary factors. By independently optimizing function-calling, retrieval, and response generation, we've demonstrated superior performance compared to fixed-pipeline systems that obscure these critical distinctions.

\section{Acknowledgements}

This work is supported by the National Science Foundation Graduate Research Fellowship Program under NSF grant number: DGE2140739. Any opinions, findings, and conclusions or recommendations expressed in this material are those of the author(s) and do not necessarily reflect the views of the National Science Foundation.

\section{Limitations}
Our work focuses on the quality of automated responses rather than direct impacts on learning outcomes, which should be explored in future work. Our evaluation scope is limited to a single large-enrollment computing course, which enabled consistent annotation quality but may not fully generalize to other disciplines, smaller discussion-based courses, or less structured educational contexts. Course-specific factors—such as programming-focused assignments and the EdSTEM platform—may further limit transferability. 

Our structure-aware retrieval approach assumes well-organized course materials with clear hierarchical structure, which may not exist in all educational settings. Additionally, offline evaluation limits assessment of real-time effectiveness. Because several experiments use proprietary LLMs, exact replication may be constrained by API changes or access restrictions. To enhance transparency, we release all prompts, dataset splits, and evaluation scripts for reproduction with open-weight or newer models. Future work should deploy EduMod-LLM across diverse courses and domains to evaluate broader applicability.

\section{Potential Negative Social Impact}
Automated educational assistants risk reinforcing educational inequities if well-resourced institutions implement them effectively while under-resourced schools lack necessary infrastructure. Over-reliance on AI assistance could discourage deeper engagement with materials, with students optimizing for quick answers rather than developing understanding. Privacy concerns arise from storing and analyzing student questions, potentially creating misusable records of student struggles without proper governance. Finally, optimizing for automated metrics could produce systems that appear effective computationally but fail to support genuine learning objectives.



\bibliography{aaai2026}

\appendix

\section{Additional Implementation Details}
\label{supp: additional_impl_details}
\subsection{Compute Costs}
Our model deployments were hosted in the cloud through Azure. Therefore, our setup did not incur any additional compute costs such as high-performance GPUs.

\subsection{LLM-as-a-Judge Prompt}

To provide more clarity on our approach, we provide the exact prompt we used for our LLM-as-a-Judge module, with few-shot examples removed to protect student privacy:

\lstset{
basicstyle=\small\ttfamily,
columns=flexible,
breaklines=true
}
\begin{lstlisting}
"""You are an expert at grading responses to student questions. You are given:
- a student question
- an LLM-written answer
- a TA-written ground-truth answer.
Assign a score from 1 to 5 for factuality and relevance:
1. Factuality: Evaluates the correctness of the information provided in the LLM-written response.
2. Relevance:  Evaluates the degree to which the LLM-written response is pertinent or related to the given student question and course.
And assign a score from 1 to 3 for style:
1. Style: Evaluates the degree to which the coherence, length, and the use of solutions, hints, and examples in the LLM-written response are appropriate for the given student question.
Please refer to the Ground Truth answer as the gold standard for all of the metrics.
Respond ONLY in dictionary format like this:
{"factuality": <1-5>, "relevance": <1-5>, "style": <1-3>}
Do NOT use the tag "json" in the response, or any backticks.
You are a kind grader. If you are ever deciding between 2 scores, choose the higher one."""

\end{lstlisting}

\subsection{Function-Calling Prompt}

We also provide the prompt used for function-calling and response generation:

\lstset{
basicstyle=\small\ttfamily,
columns=flexible,
breaklines=true
}
\begin{lstlisting}
"""You will simulate the role of a teaching assistant for an undergraduate data science course, answering student questions on a course discussion forum.
1. Your responses should be clear, helpful, and maintain a positive tone. 
2. Aim for conciseness and clarity. 
3. Use the excerpts from any solutions, course notes, and historical question-answer pairs provided to you as your primary source of information. 
4. If the question is difficult to answer based on the provided context, reply, 'Sorry, I do not know. Please wait for a staff member's response."""
\end{lstlisting}


\section{Human Evaluation Details}
\label{appendix:sec:human_eval_details}
We conducted a human evaluation of 180 model responses with two expert teaching assistants (TAs) who were blinded to the model identities. All TAs at {\university} are strong students selected and interviewed for the role and also required to take a pedagogy course. For high-quality feedback and evaluation, expert TAs with multiple years of experience were chosen for this study (one MS student and one PhD student).

To establish inter-rater reliability (IRR), we used a set of 20 representative questions. The TAs rated the factuality (1-5), relevance (1-5) and style (1-3) of each of the responses, and then measured exact match agreement. The style score was decided to be done on a scale of 1-3 to limit the inherent subjectivity of the metric. After an initial independent round of scoring, one round of discussion and a second round of evaluation, the raters achieved over 70\% alignment in each category (85\% Factuality agreement, 80\% Relevance agreement, 75\% Style agreement). We use direct percent agreement as our alignment metric.

A similar human evaluation was used to determine the optimal function calls for a set of 60 questions. Two expert TAs labeled each question with the functions they found to be most relevant, and a third TA independently resolved any discrepancies in the initial scoring.

The reviewers are all expert TAs from the US-based university where the data was collected. We employed experienced TAs rather than instructors for annotation and calibration 
because TAs are the primary responders to student questions on the discussion forum and therefore most familiar 
with the expected level of detail, tone, and pedagogical norms of the course. Instructor time constraints and differing granularity of instructor feedback also made TA annotation more practical and representative 
of real classroom operations.All persons are based in the USA.

\section{Data Privacy and Collection Details}
We understand that our data set incorporated personally identifiable information (PII) from students through their inquiries, so we took the utmost care to ensure responsible data handling and received the necessary IRB approvals for data collection and human study with TA involvement. When asking questions on the Ed student discussion board, students have the option to give permission to send their inquiry to our Azure deployment to retrieve an LLM response. We process student questions through the LLM system if and only if students explicitly give permission in their initial inquiry.  TAs screened questions for appropriateness (removing personally identifiable information and extension requests) before triggering {\model}, which generated responses only visible to TAs of the course. After review and edits, the TAs made the response visible to the students. As this is private data, there is no relevant license.


\section{EduMod-LLM Synthetic Examples}
\label{appendix:sec:synthetic_examples}
\subsection{TA-In-The-Loop Synthetic Examples}

We adopt an intuitive two-step approach to incorporating TA feedback when automatically answering student questions. After the student provides consent to using EduMod-LLM and asks a question, the bot provides a response that is only visible to the TA approved to use the bot. The TA then carefully reviews the response, and does one of 3 things: (1) leaves the response unchanged, (2) edits the response in place, or (3) completely rewrites the response. We provide some categories of edits that TAs tend to make to a response using some completely synthetic but structurally representative examples below:

\begin{tcolorbox}[title=Reducing Response Length for Conciseness, colback=blue!5, colframe=blue!40!black]
\textbf{Student Question:} ``My pandas DataFrame isn't loading correctly from the CSV file for Assignment 3. I keep getting a parsing error on line 47.''

\vspace{0.2cm}
\textbf{Original LLM Response:} ``There are several potential issues that could cause this parsing error. First, check if there are any special characters or encoding issues in your CSV file - you might need to specify the encoding parameter like \texttt{pd.read\_csv('file.csv', encoding='utf-8')} or try \texttt{'latin-1'} if that doesn't work. Second, there could be inconsistent delimiters or quote characters, so try using \texttt{pd.read\_csv('file.csv', sep=',', quotechar='"')} with explicit parameters. Third, check your system's available memory and pandas version compatibility, as older versions might handle large files differently. You should also verify that your Python environment has sufficient disk space and that no other processes are accessing the file simultaneously.''

\vspace{0.2cm}
\textbf{TA-Edited Response:} ``The parsing error is likely due to a malformed row in your CSV. Try using \texttt{pd.read\_csv('file.csv', on\_bad\_lines='skip')} to skip problematic lines and identify which row is causing the issue.''
\end{tcolorbox}

\begin{tcolorbox}[title=Editing a Detail to Improve Factuality, colback=green!5, colframe=green!40!black]
\textbf{Student Question:} ``I'm trying to calculate the correlation coefficient between two variables in my dataset. What's the difference between Pearson and Spearman correlation?''

\vspace{0.2cm}
\textbf{Original LLM Response:} ``Pearson correlation measures linear relationships between continuous variables and assumes both variables are normally distributed. Spearman correlation measures monotonic relationships and works with ordinal data. Pearson values range from -1 to 1, while Spearman values range from 0 to 1.''

\vspace{0.2cm}
\textbf{TA-Edited Response:} ``Pearson correlation measures linear relationships between continuous variables and assumes both variables are normally distributed. Spearman correlation measures monotonic relationships and works with ordinal data. Both Pearson and Spearman values range from -1 to 1, where -1 indicates perfect negative correlation and +1 indicates perfect positive correlation.''
\end{tcolorbox}

\begin{tcolorbox}[title=Rewriting an Irrelevant Response Completely, colback=orange!5, colframe=orange!40!black]
\textbf{Student Question:} ``I can't submit my Jupyter notebook for Assignment 2. The submit button isn't working.''

\vspace{0.2cm}
\textbf{Original LLM Response:} ``This submission issue could be caused by several factors. First, try refreshing your browser and clearing your cache. Make sure you're using a supported browser like Chrome or Firefox. Check your internet connection stability and try disabling browser extensions that might interfere with the submission process. You should also verify that your notebook file isn't corrupted by opening it in a new tab. If the problem persists, try restarting your Jupyter server, updating your Jupyter installation, or switching to a different network connection.''

\vspace{0.2cm}
\textbf{TA-Edited Response:} ``[Course-Specific Autograder Solution]''
\end{tcolorbox}

\end{document}